\documentclass{llncs}
\usepackage{times}
\usepackage{latexsym} 
\usepackage{graphicx}
\usepackage{named}

\input{epsf}
\newcommand{\myOmit}[1]{}
\newtheorem{mytheorem}{Proposition}

\title{Online cake cutting}

\author{Toby Walsh}

\institute{NICTA and UNSW Sydney, Australia\\ 
Email: Toby.Walsh@nicta.com.au}

\begin{document}

\maketitle

\newcommand{\mynewproof}{\noindent {\bf Proof:\ \ }}
\newcommand{\myqed}{\mbox{$\Box$}}
\newcommand{\nmax}{N}

\begin{abstract}
We propose an online form of the 
cake cutting problem. 
This models situations where
agents arrive and depart during the
process of dividing a resource. 
We show that well known fair division
procedures like cut-and-choose and
the Dubins-Spanier moving knife procedure can be 
adapted to apply to such online
problems. We propose some
fairness properties that online
cake cutting procedures can possess
like online forms of proportionality
and envy-freeness. We also consider 
the impact of collusion between 
agents. Finally, we study theoretically
and empirically the competitive ratio
of these online cake cutting procedures. 
Based on its resistance to collusion, and
its good performance in practice, our
results favour the online version of the
cut-and-choose procedure over the online
version of the moving knife procedure. 
\end{abstract}

\section{Introduction}

\begin{quote}
{\em Congratulations. Today is your birthday so you
take a cake into the office to share with your
colleagues. At tea time, people slowly start to arrive.
However, as some people have to leave early, 
you cannot wait for everyone to arrive
before you start sharing the cake.  How 
do you proceed fairly? }
\end{quote}

This is an example of an {\em online}
cake cutting problem. Most previous studies of
fair division assume that all agents
are available at the time of the division
\cite{cakecut}. Here, 
agents arrive and depart as
the cake is being divided. Online cake cutting 
provides an abstract model for a range of practical problems 
besides birthday parties. Consider, for instance,
allocating time on a large telescope. 
Astronomers will have different preferences
for when to use the 
telescope depending on what objects
are visible, the position of the sun, etc.
How do we design a web-based reservation system 
so that astronomers can asynchronously choose observation
times that is fair to all?
As a second example, consider allocating
space at an exhibition. 
Exhibitors will have different preferences
for space depending on the size, 
location, cost, etc.
How do allocate space when
not all exhibitors arrive at the 
same time but those who have arrived want to
start setting up immediately? 

Online cake cutting poses some interesting new challenges. 
On the one hand, the online aspect of such problems makes
fair division more difficult than in
the offline case. How can we 
ensure that agents do not envy
cake already given to other agents?
On the
other hand, the online aspect of such 
problems may make fair division easier
than in the offline case. Perhaps
agents do not envy cake that has
already been eaten before they arrive?

\section{Online cake cutting}

We assume that agents are risk
averse. That is, they declare valuations of the
cake that maximizes
the minimum value of the 
cake that they receive, regardless of
what the other agents do. 
This is a common assumption in cake cutting.
For instance, Brams, Jones and Klamler
(\citeyear{bjkams2006}) argue:
\begin{quote}
{\em
``\ldots As is usual in the cake-cutting literature, we postulate that the goal of each person is to maximize the value of the minimum-size piece (maximin piece) that he or she can guarantee, regardless of what the other person does. Thus, we assume that each person is risk-averse: He or she will never choose a strategy that may yield a more valuable piece of cake if it entails the possibility of getting less than a maximin piece \ldots''
}
\end{quote}
We will formulate cake cutting 
as dividing the unit interval, $[0,1]$ between
$n$ agents. 

\begin{definition}[Cutting]
A {\em cutting} of a set of intervals $S$
is a set of intervals recursively defined as follows:
$S$ is a cutting, and 
if $S$ is a cutting and
$[a,b] \in S$ then $S \setminus \{[a,b]\} \cup \{ [a,c],[c,b]\}$
is a cutting where $a < c < b$. 
\end{definition}

A partition of a set $S$ is a set of subsets of $S$ whose
union equals the original set which have an
empty pairwise intersection. That is, $\{ S_i \ | \ 1 \leq i \leq n \}$ is 
a partition of $S$ iff $S = \bigcup_{1 \leq i \leq n} S_i$ and
$S_i \cap S_j = \{\}$ for $1 \leq i < j \leq n$. 

\begin{definition}[Division]
A {\em division} of 
the cake amongst $n$ agents is a partition of some cutting 
of $\{[0,1]\}$ into $n$ subsets.
\end{definition}

A special type of division is where each agent receives a single
continuous interval. That is, the cutting contains $n$
intervals, and each agent receives a subset containing
just one interval. Note that we suppose there is no
waste and that all cake is allocated. We can
either relax this assumption, or introduce an 
additional ``dummy'' agent who 
is allocated any remaining cake. 

Agents may value parts of the cake
differently. For instance, one may prefer
the iced part, whilst another
prefers the candied part. As a second example,
as we argued before, astronomers may prefer different 
observation times.  We capture these differences
by means of valuation functions on intervals.

\begin{definition}[Valuation]
Each agent $i$ has an additive (but possibly different) valuation 
function with $v_i([0,1])=1$, 
$v_i([a,b])=v_i([a,c])+v_i([c,b])$ for any $a \leq c \leq b$,
and for some set of intervals $S$,
$v_i(S)=\sum_{ [a,b]\in S} v_i([a,b])$. 
\end{definition}

In an online cake cutting 
problem, the agents are assumed to arrive in a
fixed order. We assume without
loss of generality that the arrival order is agent 1 to agent $n$. 
Once agents are allocated all their cake, they 
depart.  The order in which agents 
are allocated cake and depart 
depends on the cake cutting
procedure. For example, the agent present
who most values the next slice of cake
could be the next to be allocated cake and
to depart. We can now formally define
the online cake cutting problem.

\begin{definition}[Online cake cutting]
An online cake cutting procedure is
a procedure that given the total
number of agents yet to arrive, 
a set of agents currently present, 
and a set of intervals $R$, 
either returns ``wait'' (indicating
that we will wait for the next agent to
arrive) or returns an agent from amongst those present
and two sets of intervals $S$ and $T$ such 
that $S \cup T$ is a cutting of $R$.
The agent returned by the procedure is allocated
$S$, and $T$ is then left to be divided amongst
the agents not yet allocated cake. 
When no agents are left to arrive
and there is only one agent present, 
the procedure must return $S=R$, $T=\{\}$.
That is, the last agent is allocated whatever
is left of the cake.
When no agents are left to arrive
and there is more than one agent 
present, the procedure cannot return ``wait''
but must cut the cake and assign it to
one agent. 
\end{definition}

Our definition of online 
cake cutting does not assume that all agents receive 
cake. Any agent can be allocated an empty set
containing no intervals. However, our definition does assume
the whole cake is eventually allocated,
and that each
agent receives all their cake at one time. 
We assume that at least one agent
is allocated some cake before the last arrives otherwise
the problem is not online.
A special type of online cake cutting procedure
is when the departure order is 
fixed in advance. For instance, 
if the procedure waits for the first
agent to arrive, and whenever a new agent
arrives, allocates cake to the
longest waiting agent then the 
departure order is the same as the arrival
order. Another special type of 
cake online cake cutting procedure
is one in which the cake is only cut from 
one or other of the ends of the cake. 
There are many interesting possible generalisations
of this problem. For example, 
there may only be a bound on the
total number of agents to arrive (e.g. you've invited
20 work colleagues to share your birthday cake
but not all of them might turn up).
Another generalisation is when an agent 
is not allocated cake all at one time but
at several moments during
the process of division. 

\section{Fairness properties}

What properties do we want from an
online cake cutting procedure? 
The literature on cake cutting studies
various notions of fairness like
envy freeness, as well
as various forms of
strategy proofness
\cite{cakecut,cakecut2,clppaaai2010}. 
These are all properties that we
might want from an online cake
cutting procedure.

\begin{description}
\item[Proportionality:]
A cake cutting procedure is
\emph{proportional} iff each of the $n$ agents assigns
at least $\frac{1}{n}$ of the total value to their piece(s).
We call such an allocation {\em proportional}. 
\item[Envy freeness:]
This is a stronger notion of fairness. 
A cake cutting procedure is 
\emph{envy free} iff no agent values another
agent's pieces more than their own.
Note that envy freeness implies proportionality
but not vice versa. 
\item[Equitability:]
A cake cutting procedure is \emph{equitable} iff
agents assign the same value to the cake
which they are allocated 
(and so no agent envies the valuation that another
agent gives to their cake). 
For 3 or more agent, equitability and
envy freeness can be incompatible
\cite{cakecut}. 
\item[Efficiency:]
This is also called Pareto optimality.
A cake cutting procedure is \emph{Pareto optimal} iff 
there is no other allocation to the one returned
that is more valuable for one agent
and at least as valuable for the others. 
Note that Pareto optimality does not in itself
ensure fairness since allocating
all the cake to one agent is Pareto optimal. 
A cake cutting procedure is \emph{weakly Pareto optimal}
iff there is no
other allocation to the one returned
that is more valuable for all agents.
A cake cutting procedure that is Pareto optimal is weakly
Pareto optimal but not vice versa.
\item[Truthfulness:]
Another consideration is whether agents
can profit by being untruthful about
their valuations. 
As in \cite{clppaaai2010},
we say that a cake cutting procedure is \emph{weakly truthful} 
iff there exists some 
valuations of the other agents such that
an agent will do at least as well by telling the truth. 
A stronger notion (often called strategy proofness
in social choice) is that agents must not be able to profit 
even when they know how others value the cake. 
As in \cite{clppaaai2010}, we say that
a cake cutting procedure is \emph{truthful} iff 
there are no valuations where an agent will do
better by lying. 
\end{description}

The fact that some agents may depart
before others arrive places some fundamental
limitations on the fairness of online cake
cutting procedures. In particular, unlike
the offline case, we can prove a strong
impossibility result.

\begin{mytheorem}
No online cake cutting procedure
is proportional, envy free or equitable.
\end{mytheorem}
\mynewproof
Consider any cake cutting procedure.
As the 
procedure is online,
at least one agent $i$ departs before 
the final agent $n$ arrives.
Since the valuation function of agent $n$, $v_n$ is not
revealed before agent $i$ 
departs, the set of intervals $S_i$ allocated to
agent $i$ cannot depend
on $v_n$. Similarly,
$v_n$ cannot change who is first to depart. 
Suppose
agent $n$ 
has a valuation function with
$v_n(S_i)=1$. 
As $v_n$ is additive and $v_n([0,1])=1$, 
agent $n$ only assigns value to the intervals
assigned to agent $i$. 
Hence, any interval outside $S_i$ 
that is allocated to agent $n$ 
is of no value to agent $n$. Hence the 
procedure is not proportional. 
Since envy-freeness implies proportionality,
by modus tollens, the procedure is also not
envy-free. 

To demonstrate that no cake cutting procedure 
is equitable, we restrict ourselves to problems 
in which all agents assign non-zero value
to any non-empty interval. 
Suppose that the procedure is equitable.
As all the cake is allocated, at least
one agent must receive cake. Since
the procedure is equitable, it follows
that all agents must receive some cake. 
Now, the first agent $i$ to depart
and the set of intervals $S_i$ allocated to agent $i$ 
cannot depend
on $v_n$, the valuation function of
the last agent to arrive.  Suppose $v_i(S_i)=a$. 
Now we have argued that $S_i$ is non-empty. 
Hence, by assumption, $a>0$. 
We now modify the valuation
function of agent $n$
so that $v_n(S_i)=1-\frac{a}{2}$. 
Then $v_n(S_n) \leq \frac{a}{2} < a = v_i(S_i)$. 
Hence
the procedure is not equitable. 
\myqed

By comparison, 
the other properties of Pareto optimality and
truthfulness are achievable in the online
setting. 

\begin{mytheorem}
There exist online cake cutting procedures
that are Pareto optimal and truthful.
\end{mytheorem}
\mynewproof
Consider the online cake cutting procedure
which allocates all cake to the first agent
to arrive. This is Pareto optimal as any
other allocation will be less desirable for
this agent. It is also truthful as no
agent can profit by lying about their 
valuations. 
\myqed

Of course, allocating all cake to the 
first agent to arrive is not a very
``fair'' procedure. Therefore we need to consider
other weaker properties of fairness 
that online procedures can possess. 
We introduce such properties in the next 
section

\section{Online properties}

We define some fairness properties that are
specific to online procedures. 
\begin{description}
\item[Proportionality:]
We weaken the definition of
proportionality to test whether agents receive a fair proportion
of the cake that {\em remains} when they arrive.
A cake cutting procedure is
\emph{weakly proportional} iff 
each agent assigns
at least $\frac{r}{k}$ of the total value 
of the cake to their pieces
where $r$ is the fraction of the total
value assigned by the agent to the
(remaining) cake when they arrive and $k$ is the number of
agents yet to be allocated cake at this point.
\item[Envy freeness:]
We can weaken the definition of envy freeness
to consider just agents allocated cake after
the arrival of a given agent. 
A cake cutting procedure is
\emph{weakly envy free} iff agents do not value
cake allocated to agents after
their arrival more than their own. 
Note that weak envy freeness implies weak proportionality
but not vice versa. Similarly,
envy freeness implies weak envy freeness but
not vice versa.  
An even weaker form of envy freeness
is when an agent only envies cake 
allocated to other agents whilst they are present.
A cake cutting procedure is
\emph{immediately envy free} iff agents do not value
cake allocated to any agent after
their arrival and before their departure more than their own. 
Weak envy freeness implies immediate envy freeness
but not vice versa. 
\item[Order monotonicity:]
An agent's allocation of cake
typically depends on when they
arrive. We say that 
a cake cutting procedure is
\emph{order monotonic} iff
an agent's valuation of their
cake does not decrease when
they are moved earlier in the
arrival ordering and all other agents
are left in the same relative positions.
Note that as the moved agent
can receive cake of greater
value, other agents
may receive cake of less value. 
A positive interpretation of order monotonicity
is that agents are encouraged
to participate as early as possible.
On the other hand, order monotonicity 
also means that agents who
have to arrive late due to reasons 
beyond their control may
receive less value. 
\end{description}

The online versions of the 
proportional and envy free properties are weaker
than their corresponding offline
properties. 
We consider next two well known
offline procedures that naturally
adapt to the online setting and demonstrate
that they have many of the 
online properties 
introduced here. 

\section{Online Cut-and-Choose}

The cut-and-choose procedure for two
agents dates back to antiquity. 
It appears nearly three thousand
years ago in Hesiod's poem {\em Theogeny}
where Prometheus
divides a cow and Zeus selects the
part he prefers. 
Cut-and-choose is enshrined in the UN's 1982 Convention of the Law of the Sea
where it is used to divide the seabed for mining. 
In cut-and-choose, one agent
cuts the cake and the other takes
the ``half'' that they most prefer. 
We can extend cut-and-choose to more than two
agents by having one agent cut a ``proportional''
slice and giving this slice to the agent who values it
most. We then repeat with one
fewer agent. The two person cut-and-choose procedure
is proportional, envy free, Pareto optimal and 
weakly truthful. 
However, it
is not equitable nor truthful. 

We can use cut-and-choose as the
basis of an online cake cutting procedure. 
The first agent to arrive cuts
off a slice of cake and waits for the next
agent to arrive. Either the next agent to arrive
chooses this slice and departs, or the
next agent to arrive declines this slice
and the waiting agent takes this slice and departs.
If more agents are to arrive, the remaining
agent cuts the cake and we repeat the process.
Otherwise, the remaining agent is the
last agent to be allocated cake and departs
with whatever is left. 
We assume that all agents 
know how many agents will arrive. 
A natural extension (which we do not
consider further) is when 
multiple agents arrive and can choose
or reject the cut cake. 
By insisting that an agent cuts the
cake before the next agent is allowed
to arrive, we will make the procedure
more resistant to collusion. We discuss
this in more detail later. 

\begin{example}
Suppose there are three agents,
the first 
values only $[\frac{1}{2},1]$,
the second 
values only $[\frac{1}{3},1]$,
and the third 
values only $[0,\frac{3}{4}]$.
We suppose that they uniformly
value slices within these intervals.
If we operate the online 
cut-and-choose procedure, the first
agent arrives and cuts off the
slice $[0,\frac{2}{3}]$ as they
assign this slice $\frac{1}{3}$ the total
value of the cake. 
The second agent then arrives. 
As they assign this
slice with $\frac{1}{2}$ the total value of
the cake and they are only expecting
$\frac{1}{3}$ of the total, the second agent is 
happy to take this slice and
depart. The first agent then cuts off the
slice $[\frac{2}{3},\frac{5}{6}]$ as 
they assign this $\frac{1}{3}$
of the total value of the cake (and $\frac{1}{2}$
of the value remaining after the second
agent departed with their slice). 
The third agent then arrives.
As they assign the slice 
$[\frac{2}{3},\frac{5}{6}]$ 
with all of the 
total value of the remaining cake and
they are only expecting $\frac{1}{2}$ of 
whatever remains, the
third agent is happy to
take this slice and depart.
The first agent now takes
what remains, the slice 
$[\frac{5}{6},1]$. 
We can argue that everyone 
is happy as the first agent received a ``fair''
proportion of the 
cake, whilst the other two agents
received slices that were of even greater proportional
value to them. 
\end{example}

The online cut-and-choose
procedure has almost all of the
online fairness properties just introduced.

\begin{mytheorem}
The online cut-and-choose procedure
is weakly proportional, immediately envy free,
and weakly truthful. 
However, it is not
proportional, (weakly) envy free, equitable,
(weakly) Pareto optimal, truthful or order monotonic. 
\end{mytheorem}
\mynewproof
Suppose agent $i$ cuts the
slice $c_i$. As agent $i$ is risk averse,
and as there is a chance that agent $i$
is allocated $c_i$, agent $i$ will cut $c_i$ 
to ensure that $v_i(c_i) \geq \frac{r}{k}$ where 
$k$ is the number of agents still to
be allocated cake and $r$ is the fraction of 
cake remaining when agent $i$ arrived. Similarly
as there is a chance that agent $i$
is not allocated $c_i$, but will
have to take a share of what remains, 
they will cut $c_i$ so that $v_i(c_i) \leq \frac{r}{k}$. 
Hence, $v_i(c_i) = \frac{r}{k}$,
and the procedure is both weakly proportional
and weakly truthful. 
It is also immediately envy free
since each slice that agent $i$ cuts
(and sees allocated) has the same value, $\frac{r}{k}$. 
%
\myOmit{
To demonstrate surjectivity,
consider the partition that 
allocates the $i$th agent with
the slice $[a_i,a_{i+1}]$ 
where $a_1=0$ and $a_{n+1}=1$. 
We construct a valuation 
for the $i$th agent ($i<n-1$)
that assigns a value $0$ to $[0,a_i]$,
a value $1$ to $[a_i,a_{i+1}]$,
a value $0$ to $[a_{i+1},a_{i+2}]$,
a value $n-i$ to $[a_{i+2},1]$.
For the $n-1$th agent,
we construct a valuation function
that assigns a value $0$ to $[0,a_{n-1}]$,
and values of $1$ to both $[a_{n-1},a_n]$
and  $[a_n,1]$. 
Finally, 
we construct a valuation function
for the $n$th agent
that assigns a value $0$ to $[0,a_{n}]$,
and a value of $1$ to $[a_n,1]$. 
With these valuation
functions, the $i$th agent gets the slice $[a_i,a_{i+1}]$. }

To show that this procedure is not proportional, (weakly)
envy free, equitable, (weakly) Pareto optimal,
truthful or order monotonic consider 
four agents who value the cake as follows:
$v_1([0,\frac{1}{4}])=\frac{1}{4}$,
$v_1([\frac{1}{4},\frac{3}{4}])=\frac{1}{12}$,
$v_1([\frac{3}{4},1])=\frac{2}{3}$, 
$v_2([\frac{1}{4},\frac{1}{2}])=\frac{1}{3}$,
$v_2([\frac{1}{2},\frac{5}{8}])=\frac{2}{3}$,
$v_3([0,\frac{1}{4}])=\frac{1}{2}$,
$v_3([\frac{1}{2},\frac{5}{8}])=\frac{1}{12}$,
$v_3([\frac{5}{8},\frac{3}{4}])=\frac{1}{6}$, 
$v_3([\frac{3}{4},1])=\frac{1}{4}$,
$v_4([\frac{1}{4},\frac{1}{2}])=\frac{3}{4}$, 
$v_4([\frac{1}{2},\frac{3}{4}])=\frac{1}{12}$,
and $v_4([\frac{3}{4},1])=\frac{1}{6}$.
All other slices have zero value. For instance,
$v_2([0,\frac{1}{4}])=v_3([\frac{1}{4},\frac{1}{2}])=0$.

If we apply the online cut-and-choose
procedure, agent 1
cuts off the slice $[0,\frac{1}{4}]$ as 
$v_1([0,\frac{1}{4}])=\frac{1}{4}$ and 4 agents are
to be allocated cake.
Agent 2 places no value on this slice so 
agent 1 takes it. 
Agent 2 then cuts off the slice
$[\frac{1}{4},\frac{1}{2}]$ as 
$v_2([\frac{1}{4},\frac{1}{2}])=\frac{1}{3} v_2([\frac{1}{4},1])$ 
and 3 agents remain to be allocated cake.
Agent 3 places no value on this slice so
agent 2 takes it.
Agent 3 then cuts the
cake into two pieces of equal value: $[\frac{1}{2},\frac{3}{4}]$
and $[\frac{3}{4},1]$.
Agent 4
takes the slice $[\frac{3}{4},1]$ as it has greater
value, leaving agent 3 with
the slice $[\frac{1}{2},\frac{3}{4}]$

The procedure is not proportional as 
agent 4 receives the slice $[\frac{3}{4},1]$ but
$v_4([\frac{3}{4},1])=\frac{1}{6}$. 
The procedure is not (weakly) envy free as 
agent 1 receives the slice $[0,\frac{1}{4}]$ and
agent 4 receives the slice $[\frac{3}{4},1]$, 
but $v_1([0,\frac{1}{4}]) = \frac{1}{4}$ and
$v_1([\frac{3}{4},1]) = \frac{2}{3}$. Hence 
agent 1 envies the slice allocated to agent 4. 
The procedure is not equitable as agents receive
cake of different value.
The procedure is not (weakly) Pareto optimal
as allocating agent 1
with $[\frac{3}{4},1]$, agent 2
with $[\frac{1}{2},\frac{3}{4}]$,
agent 3 with $[0,\frac{1}{4}]$,
and agent 4 with $[\frac{1}{4},\frac{1}{2}]$
gives all agents greater 
value. 

The procedure is not truthful
as agent 2 can 
get a more valuable 
slice by misrepresenting their preferences
and cutting off
the larger slice $[\frac{1}{4},\frac{5}{8}]$.
This slice contains all the cake of any value to agent 2.
Agent 3 has $v_3([\frac{1}{4},\frac{5}{8}])=\frac{1}{12}$ so
lets agent 2 take this larger slice. 
Finally, the procedure is not order
monotonic as the value of the cake
allocated to agent 4
decreases from $\frac{1}{6}$ to $\frac{1}{8}$
when they arrive
before agent 3. 
\myqed

\section{Online moving knife}

Another class of cake cutting procedures
uses one or more moving knives. 
For example, in the Dubins-Spanier procedure
for $n$ agents 
\cite{dsam61},  a knife is moved across
the cake from left to right. 
When an agent shouts ``stop'',
the cake is cut and this agent
takes the piece to the left of the knife.
The procedure continues with the remaining
agents until one agent is left
(who takes whatever remains). 
This procedure is proportional but is not
envy-free. However, only the first $n-2$ 
agents allocated slices of cake
can be envious. 

We can use the Dubins-Spanier procedure as the
basis of an online moving knife procedure. 
The first $k$ agents ($k \geq 2$) to arrive perform
one round of a moving knife procedure to select
a slice of the cake. Whoever chooses
this slice, departs. At this point, if all 
agents have arrived, we continue the
moving knife procedure with $k-1$
agents. Alternatively the next
agent arrives and we start again 
a moving knife procedure with $k$ agents. 

\begin{example}
Consider again the example in which 
there are three agents,
the first 
values only $[\frac{1}{2},1]$,
the second 
values only $[\frac{1}{3},1]$,
and the third 
values only $[0,\frac{3}{4}]$.
If we operate the online 
moving knife procedure, the first
two agents arrive and 
perform one round of the moving
knife procedure. 
The second agent 
is the first to call ``cut'' 
and departs with the 
slice $[0,\frac{5}{9}]$ (as this
has $\frac{1}{3}$ of the
total value of the cake for them).
The third agent 
then arrives and performs a 
round of the moving knife
procedure with the first
agent using the remaining cake. 
The third agent is the
first to call ``cut'' and
departs with the slice
$[\frac{5}{9},\frac{47}{72}]$
(as this has $\frac{1}{2}$ 
the total value of the remaining
cake for them). 
The first agent 
takes what remains, the slice
$[\frac{47}{72},1]$. 
We can argue that everyone 
is happy as the second and third
agents received a ``fair''' proportion of the 
cake that was left when they arrived, whilst
the first agent received an even greater 
proportional value. 
\end{example}

The online moving knife
procedure has similar fairness properties 
as the online cut-and-choose
procedure. However, as we shall show in the
following sections, it is neither as resistant
to collusion nor as fair in practice. 

\begin{mytheorem}
The online moving knife procedure
is weakly proportional, immediately envy free
and weakly truthful. 
However, it is not
proportional, (weakly) envy free, equitable, 
(weakly) Pareto optimal, truthful or order monotonic. 
\end{mytheorem}
\mynewproof
Suppose $j$ agents ($j>1$) have still
to be allocated cake. Consider
any agent who has arrived. They 
call ``cut'' as soon
as the knife reaches $\frac{1}{j}$ of the value 
of the cake left for fear that
they will receive cake of 
less value at a later stage. 
Hence, the procedure is
weakly truthful and weakly proportional.
The procedure is also immediately
envy free as they will assign less value 
to any slice that is allocated
after their arrival and before their
departure. 
\myOmit{
To demonstrate surjectivity,
consider the partition that 
allocates the $i$th agent with
the slice $[a_i,a_{i+1}]$ 
where $a_1=0$ and $a_{n+1}=1$. 
We construct a valuation 
for the $i$th agent ($i<n$)
that assigns a value $0$ to $[0,a_i]$,
a value $1$ to $[a_i,a_{i+1}]$,
a value $n-i$ to $[a_{i+1},1]$.
Finally, 
we construct a valuation function
for the $n$th agent
that assigns a value $0$ to $[0,a_{n}]$,
and a value of $1$ to $[a_n,1]$. 
With these valuation
functions, the $i$th agent gets the slice $[a_i,a_{i+1}]$. 
}

To show that this procedure is not proportional, (weakly)
envy free, equitable, (weakly) Pareto optimal, or
truthful consider
again the example with four agents used
in the last proof. Suppose $k=2$ so that
two agents perform each round of the moving
knife procedure. Agent 1 and 2 arrive and run
a round of the moving knife procedure.
Agent 1 calls ``cut'' and
departs with the slice $[0,\frac{1}{4}]$.
Agent 3 then arrives and agent 2 and 3 perform
a second round of the moving knife procedure. 
Agent 2 calls ``cut'' and
departs with the slice $[\frac{1}{4},\frac{1}{2}]$.
Agent 4 then arrives and agent 3 and 4 perform
the third and final round of the moving knife
procedure. Agent 3
calls ``cut'' and 
departs with the slice $[\frac{1}{2},\frac{3}{4}]$,
leaving agent 4 with
the slice $[\frac{3}{4},1]$. This
is the same allocation as the online
cut-and-choose procedure.
Hence, for the same reasons as before,
the online moving knife procedure is not proportional,
(weakly) envy free, (weakly) Pareto optimal
or truthful. 

Finally, to show that the online
moving knife procedure is not order
monotonic consider again $k=2$, and three agents
with valuation functions:
$v_1([0,\frac{1}{3}]) = v_1([\frac{1}{3},\frac{2}{3}]) = 
v_1([\frac{2}{3},1]) =  \frac{1}{3}$,
$v_2([0,\frac{1}{3}]) = 0$, 
$v_2([\frac{1}{3},\frac{2}{3}]) = 
v_2([\frac{2}{3},1]) = \frac{1}{2}$,
$v_3([0,\frac{1}{6}])= \frac{1}{3}$,
$v_3[\frac{1}{6},\frac{1}{3}]) = 
v_3([\frac{1}{3},\frac{2}{3}]) = 0$,
and $v_3([\frac{2}{3},1]) = \frac{2}{3}$.
Agent 1 and 2 arrive and run
a round of the moving knife procedure.
Agent 1 calls ``cut'' and
departs with the slice $[0,\frac{1}{3}]$.
Agent 3 then arrives and agent 2 and 3 perform
a second and final round of the moving knife procedure. 
Agent 2 calls ``cut'' and
departs with the slice $[\frac{1}{3},\frac{2}{3}]$,
leaving agent 3 with the slice
$[\frac{2}{3},1]$. On the other hand, 
if agent 3 arrives ahead
of agent 2 then the value
of the interval allocated to agent 3
drops from $\frac{2}{3}$ to $\frac{1}{3}$. 
Hence the procedure is not order
monotonic. 
\myqed

\section{Online collusion}

An important consideration in online cake
cutting procedures is whether agents present
together in the room can collude together to 
increase the amount of cake they receive. 
We shall show that this is a property
that favours the online cut-and-choose 
procedure over the online moving knife procedure.
We say that a cake cutting procedure is 
vulnerable (resistant) to 
{\em online collusion} iff there exists (does not exist)
a protocol to which the colluding agents
can agree which increases or keeps constant the value of the cake that
each receives. We suppose that agents do not 
meet in advance so can only agree to a collusion
when they meet during cake cutting. 
We also suppose that other agents can
be present when agents are colluding. 
Note that colluding agents cannot change their
arrival order and can only indirectly
influence their departure order. The arrival
order is fixed in advance, and the departure
order is fixed by the online cake cutting
procedure. 

\subsection{Online cut-and-choose}

The online cut-and-choose procedure
is resistant to online collusion. 
Consider, for instance, the first two agents to participate.
The first agent cuts the cake before the
second agent is present (and has agreed to 
any colluding protocol). As the first agent is
risk averse, they will cut the cake proportionally
for fear that the second agent will decline to 
collude. Suppose the second agent does not assign
a proportional value to this slice. It would be risky
for the second agent to agree to any protocol in which they
accept this slice as they might assign less
value to any cake which the first agent
later offers in compensation. Similarly, suppose
the second agent assigns a proportional or greater value
to this slice. It would be risky for the second agent to agree
to any protocol in which they reject
this slice as they might assign less total value
to the slice that they are later allocated and any cake
which the first agent offers them in compensation. 
Hence, assuming that the second agent is risk
averse, the second agent will follow the usual
protocol of accepting the slice iff it is at least
proportional. A similar argument can
be given for the other agents. 

\subsection{Online moving knife}
 
On the other hand, the online moving knife
procedure is vulnerable to 
online collusion. Suppose four or more agents
are cutting a cake using the online moving knife
procedure, but the first two agents
agree to the following protocol:
\begin{enumerate}
\item Each agent will (silently) indicate when
the knife is over a slice worth $\frac{3}{4}$ of 
the total.
\item Each will only call ``stop'' 
once the knife is over a slice worth $\frac{3}{4}$ of 
the total and the other colluding agent has given their (silent) indication
that the cake is also worth as much to them;
\item Away from the eyes of the other agents, the two colluding
agents will share this slice of cake 
using a moving knife procedure. 
\end{enumerate}
Under this protocol, both agents will receive
slices that they value more than $\frac{1}{4}$ of the total.
This is better than not colluding. 
Note that it is advantageous for the agents to
agree to a protocol in which they 
call ``stop'' later than this. For example,
they could agree to call stop at $\frac{(p-1)}{p}$
of the total value for some $p>3$. 
In this way, they would receive more than
$\frac{(p-1)}{2p}$ of the total value of the 
cake (which tends
to half the total value as $p \leadsto \infty$).

\myOmit{
\subsection{Collusion in advance}

Recall that 
we have assumed that agents do not 
meet in advance so can only agree to a collusion
when they first meet during cake cutting. 
If we relax this assumption then there are more
opportunities for collusion. For instance,
the online cut-and-choose procedure now 
becomes vulnerable to collusion. 
GIVE BETTER EXAMPLE
Suppose
the first two agents meet in advance
and agree to a protocol in which the second
agent indicates how large a slice
is needed to be worth $\frac{3}{n}$ of the total,
the first agent cuts a slice that is worth at least
$\frac{3}{n}$ of the total for them and the 
second agent, the second agent
rejects this slice, 
and the first agent shares the slice
with the second agent using a cut-and-choose procedure. 
Under this protocol, both agents will receive
slices that they value more than $\frac{1}{n}$ of the total. 
}

\section{Competitive analysis}

An important tool to study online algorithms is
competitive analysis. 
We say that an online algorithm is {\em competitive} iff the ratio between
its performance and the performance of the corresponding offline algorithm
is bounded. But how do we measure the 
performance of a cake cutting algorithm?

\subsection{Egalitarian measure}

An egalitarian measure of performance would be the
reciprocal of the smallest value assigned by any agent to
their slice of cake.
We take the reciprocal so that 
the performance measure increases
as agent gets less valuable slices
of cake. 
Using such a measure of performance, neither the online
cut-and-choose nor the online moving knife procedures are competitive.
There exist examples with just 3 agents
where the competitive ratio of either online procedure is
unbounded. The problem is that
the cake left to share 
between the late arriving agents
may be of very little value to these
agents.

\subsection{Utilitarian measure}

An utilitarian measure
of performance would be the reciprocal
of the sum of the values assigned by
the agents to their slices of cake (or equivalently
the reciprocal of the mean value). 
With such a measure of performance, 
the online cut-and-choose and moving knife procedures are competitive
provided the total number of agents, $n$ is bounded.
By construction, the first agent in the online
cut-and-choose or moving knife procedure must receive cake of
value at least $\frac{1}{n}$ of the total. Hence,
the sum of the valuations is at least $\frac{1}{n}$. 
On the other hand, the sum of the valuations of the corresponding
offline algorithm cannot be more than $n$.
Hence the competitive ratio cannot be more
than $n^2$. In fact,
there exist examples where 
the ratio is $O(n^2)$. Thus the utilitarian 
competitive ratio
is bounded iff $n$ itself is bounded. 

\myOmit{
We can compare this with the competitive
ratio of the (offline) moving knife procedure 
when all $k$ agents arrive at the same time. 
By construction, each agent assigns
at least $\frac{1}{k}$ of the total value
to their cake.
It is not hard, however, to construct
problems like the one above where 
each agent assigns exactly 
$\frac{1}{k}$ of the total value
to their cake but there exists perfect
allocations where each agents can assign
the total value of the cake to their slice. 
In such a case,
the competitive ratio is $k$. Thus, the offline
moving knife problem is competitive
iff the number of agents is bounded. 
}

\section{Experimental results}

To test the performance of these procedures in 
practice, we ran some experiments in which we computed the
competitive ratio of the online moving
knife and cut-and-choose procedures compared
to their offline counterparts. 
We generated piecewise linear valuations
for each agent by dividing the cake into
$k$ random segments, and assigning a
random value to each segment, normalizing the
total value of the cake. 
It is an interesting research question
whether random valuations 
are more challenging than 
valuations which are more correlated. 
For instance, if all agents have the 
same valuation function (that is, 
if we have perfect correlation) then the online
moving knife procedure performs
identically to the offline. 
On the other hand, if the valuation functions are not
correlated, online cake cutting 
procedures can struggle to be fair especially 
when late arriving agents more greatly value the
slices of cake allocated to early
departing agents.
Results obtained uncorrelated instances
need to be interpreted with 
some care as there are many pitfalls to using
instances that are generated entirely at
random \cite{how-not-to,mpswcp98,random}.

\begin{figure}[htb]
\begin{center}
\begin{tabbing}
\includegraphics[scale=0.7]{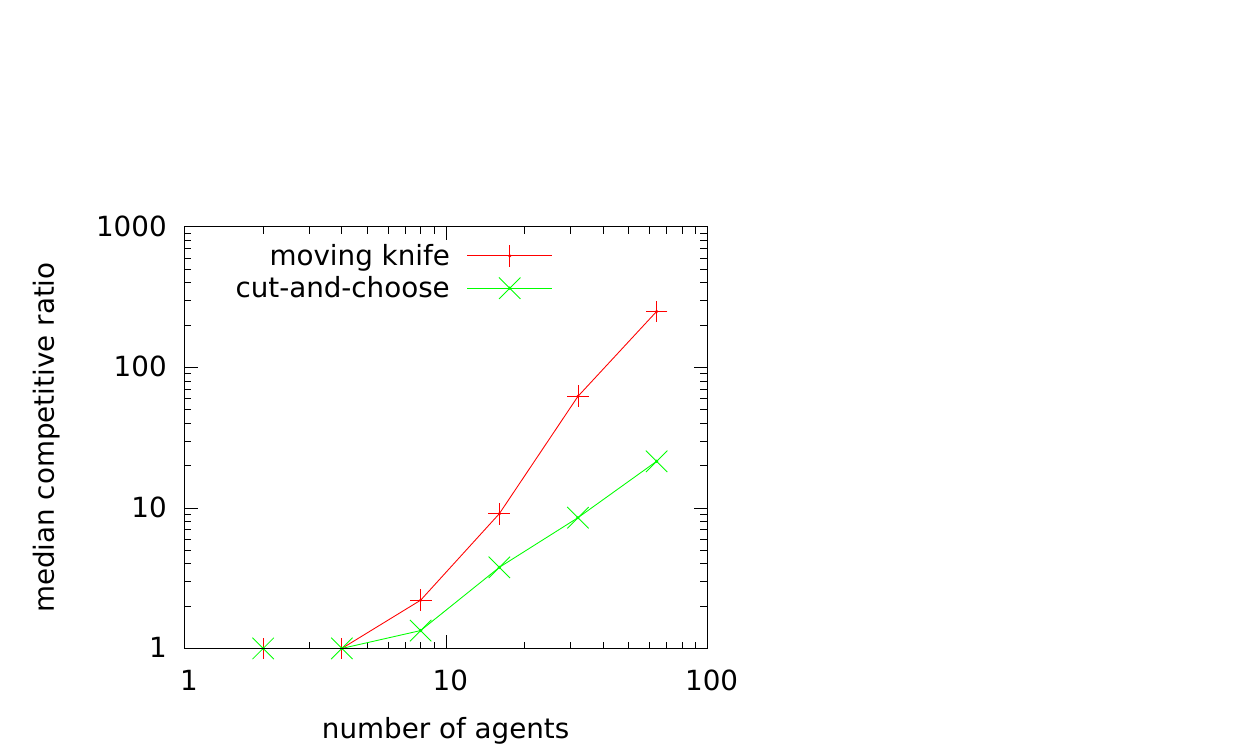} 
\hspace{-1.2in}
\=
\includegraphics[scale=0.7]{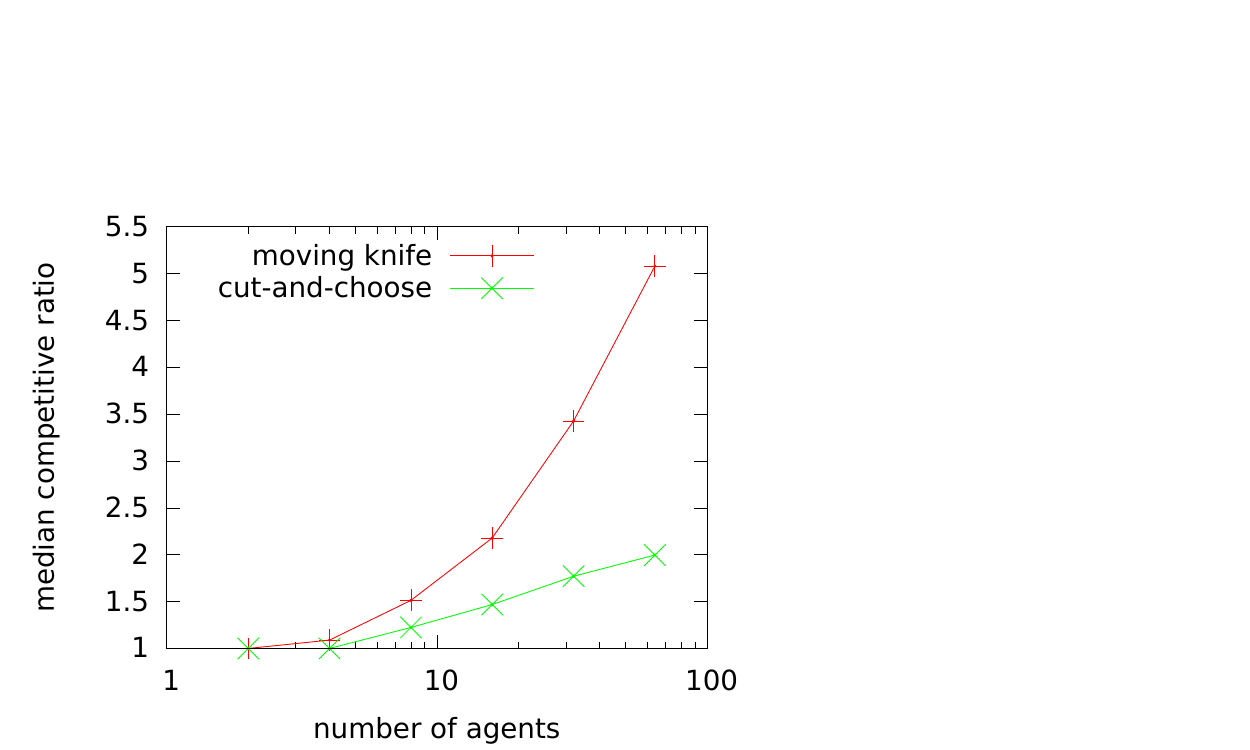} \\
\ \ \ \ \ \ \ \ \ \ \ \ \ \ \ \ \ \ \ \ \ \ \ (a) egalitarian \>  \ \ \ \ \ \ \ \ \ \ \ \ \ \ \ \ \ \ \ \ \ \ (b) utilitarian
\end{tabbing}
\end{center}
\caption{Competitive ratio between 
online and offline
cake cutting procedures for
(a) the egalitarian and (b) utilitarian performance measures. 
Note different scales to y-axes.}
\end{figure}

\myOmit{
\begin{figure}[htb]
\vspace{-0.1in}
\begin{center}
{\includegraphics{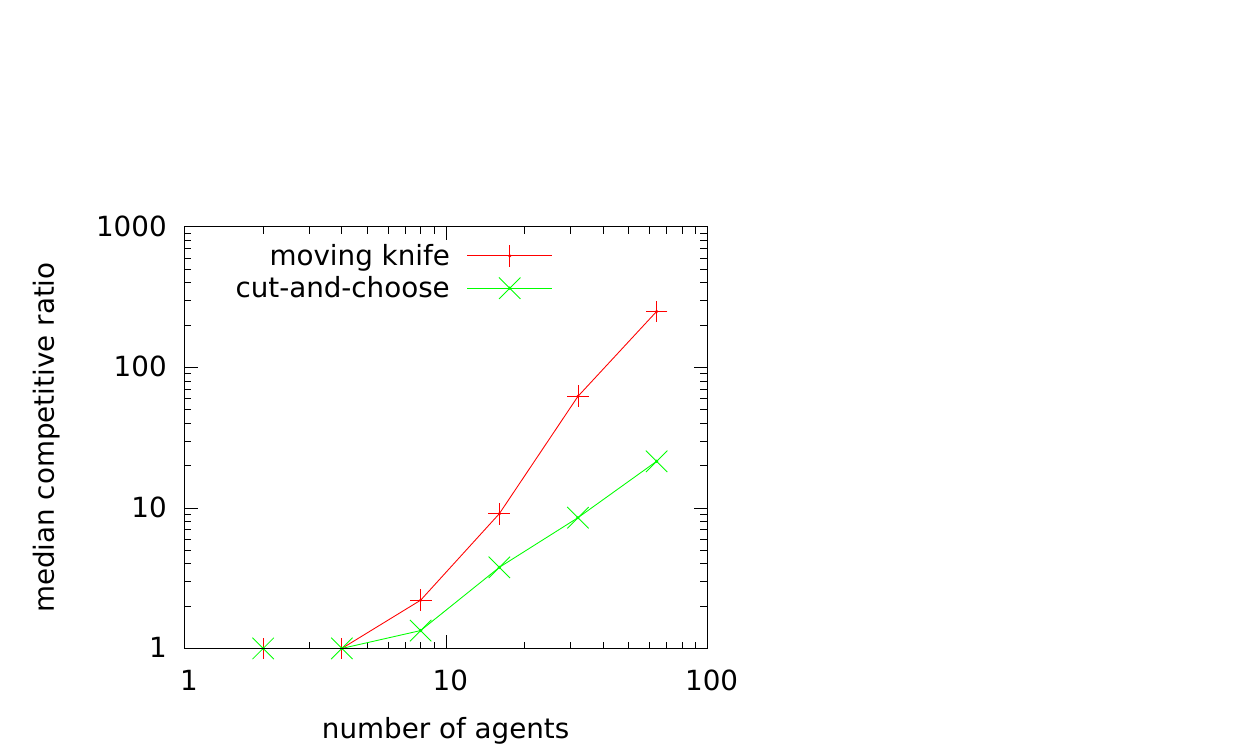}}
\end{center}
\caption{Competitive ratio between 
the egalitarian performance of online and offline
cake cutting procedures. Note log scales to both axes.}
\end{figure}

\begin{figure}[htb]
\vspace{-0.3in}
\begin{center}
{\includegraphics{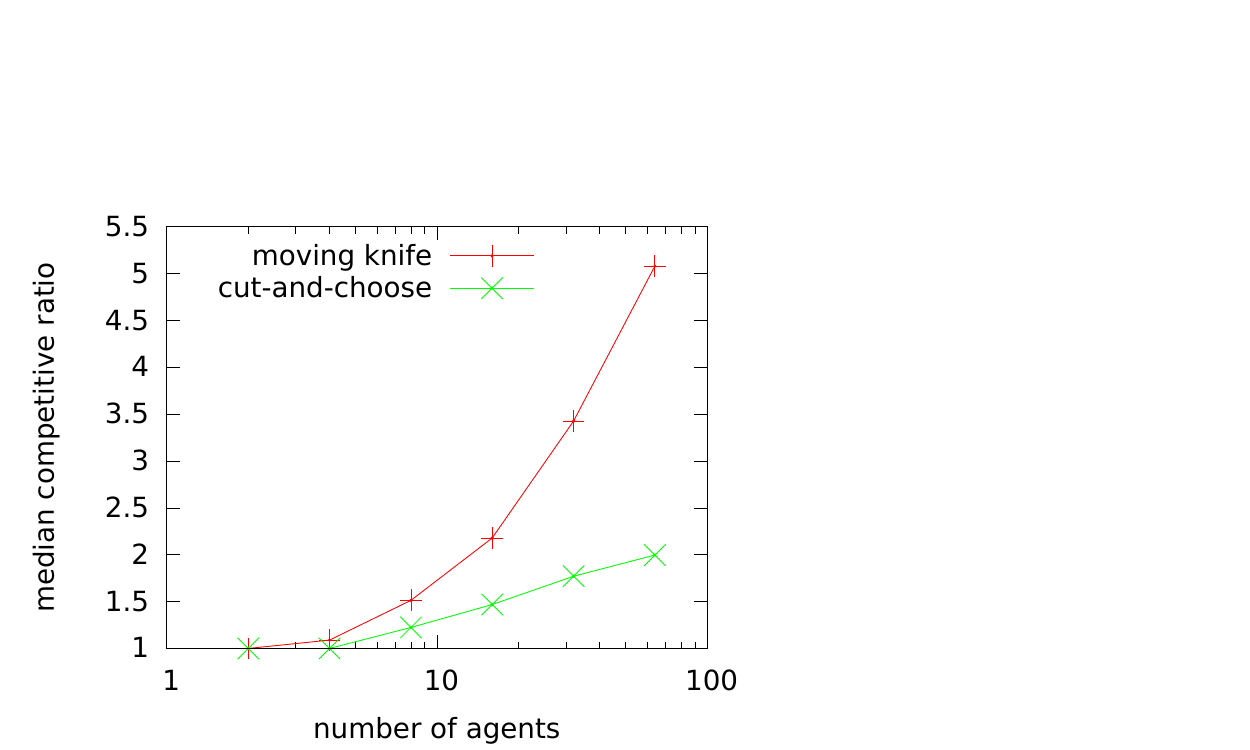}}
\end{center}
\caption{Competitive ratio between 
the utilitarian performance of online and offline
cake cutting procedures. Note linear scale to y-axis.}
\end{figure}
}

We generated cake cutting problems with between 2
and 64 agents, where each agent's valuation function
divides the cake into 8 random segments. 
At each problem size, 
we ran the online and offline moving knife and 
cut-and-choose procedures on the
same 10,000 random problems. 
Overall, the online cut-and-choose procedure 
performed much better than the online
moving knife procedure according to
both the egalitarian and utilitarian performance measures. 
By comparison, the offline moving knife procedure 
performed slightly better than the offline
cut-and-choose procedure according to
both measures. 
See Figure 1 for plots of the competitive
ratios between the performance of the online
and offline procedures. 
Perhaps unsurprisingly, the egalitarian performance
is rather disappointing when there are many agents
since there is a high probability that
one of the late arriving agents gets cake of little value.
However, the utilitarian performance is reasonable,
especially for the online cut-and-choose procedure.
With 8 agents, the 
average value of cake
assigned to an agent by the online cut-and-choose
procedure is within about 20\% of that assigned
by the offline procedure. 
Even with 64 agents, the average value 
is within a factor of 2 of that assigned
by the offline procedure. 

\section{Online Mark-and-Choose}

A possible drawback of both of the 
online cake cutting procedures proposed
so far is that the first agent to arrive
can be the last to 
depart. What if we want a 
procedure in which agents can depart
soon after they arrive? 
The next procedure has this property. 
Agents depart as soon as the next agent arrives
(except for the last agent to arrive who takes
whatever cake remains). 
However, the new procedure
may not
allocate cake from one end. 
In addition, the new procedure
does not necessarily allocate
continuous slices of cake. 

In the online mark-and-choose procedure,
the first agent
to arrive marks the cake into
$n$ pieces. The second agent
to arrive selects one piece to give
to the first agent who then 
departs. The second agent
then marks the remaining cake into 
$n-1$ pieces and waits for the
third agent to arrive. The procedure
repeats in this way until the last agent
arrives. The last agent to arrive selects
which of the two halves marked by the penultimate
agent should be allocated to the penultimate 
agent, and takes whatever remains.

\begin{example}
Consider again the example in which 
there are three agents,
the first 
values only $[\frac{1}{2},1]$,
the second 
values only $[\frac{1}{3},1]$,
and the third 
values only $[0,\frac{3}{4}]$.
If we operate the online 
mark-and-choose procedure, the first
agent arrives and 
marks the cake into 3 equally valued 
pieces:
$[0,\frac{2}{3}]$,
$[\frac{2}{3},\frac{5}{6}]$,
and 
$[\frac{5}{6},1]$.
The second agent then
arrives and selects the least valuable piece
for the first agent to take.
In fact, both 
$[\frac{2}{3},\frac{5}{6}]$
and 
$[\frac{5}{6},1]$
are each worth $\frac{1}{4}$ of the
total value of the cake to the 
second agent. The second agent
therefore chooses between them
arbitrarily. Suppose
the second agent decides to give the slice
$[\frac{2}{3},\frac{5}{6}]$
to the first agent.
Note that the first agent assigns
this slice with $\frac{1}{3}$ of the total value
of the cake. This leaves behind two sections of 
cake: $[0,\frac{2}{3}]$ and
$[\frac{5}{6},1]$.
The second agent then marks
what remains into two equally valuable
pieces: the first
is the interval $[0,\frac{7}{12}]$
and the second contains
the two intervals $[\frac{7}{12},\frac{2}{3}]$
and $[\frac{5}{6},1]$. 
The third agent then arrives
and selects the least valuable
piece for the second agent
to take. 
The first piece is worth $\frac{7}{12}$
of the total value of the cake
to the third agent. As this is over
half the total value, the other piece
must be worth less. 
In fact, the second piece is worth $\frac{1}{4}$ of
the total value. The third agent 
therefore gives the second piece
to the second agent. 
This leaves the third agent with
the remaining slice $[0,\frac{7}{12}]$. 
It can again be claimed that everyone 
is happy as the first 
agents received a ``fair''' proportion of the 
cake that was left when they arrived, whilst
both the second and third agent received an even greater 
proportional value. 
\end{example}

This procedure again has the same fairness properties 
as the online cut-and-choose
and moving knife
procedures. 

\begin{mytheorem}
The online mark-and-choose procedure
is weakly proportional, immediately envy free
and weakly truthful.  
However, it is not
proportional, (weakly) envy free, equitable, 
(weakly) Pareto optimal, truthful, or order monotonic. 
\end{mytheorem}
\mynewproof
Any agent marking the cake
divides it into slices of equal
value (for fear that they will
be allocated one of the less valuable slices). 
Similarly, an agent selecting a slice
for another agent selects
the slice of least value to them (to maximize
the value that they receive). 
Hence, the procedure is
weakly truthful and weakly proportional.
The procedure is also immediately
envy free as they will assign less value 
to the slice that they select
for the departing agent than
the value of the slices that they
mark. 
\myOmit{
To demonstrate surjectivity,
consider the partition that 
allocates the $i$th agent with
the slice $[a_i,a_{i+1}]$ 
where $a_1=0$ and $a_{n+1}=1$. 
We construct a valuation 
for the $i$th agent ($i<n$)
that assigns a value $0$ to $[0,a_i]$,
a value $1$ to $[a_i,a_{i+1}]$,
a value $n-i$ to $[a_{i+1},1]$.
Finally, 
we construct a valuation function
for the $n$th agent
that assigns a value $0$ to $[0,a_{n}]$,
and a value of $1$ to $[a_n,1]$. 
With these valuation
functions, the $i$th agent gets the slice $[a_i,a_{i+1}]$. }

To show that this procedure is not proportional, (weakly)
envy free, equitable, (weakly) Pareto optimal
or truthful consider
again the example with four agents used
in earlier proofs. 
The first agent marks and is assigned
the slice $[0,\frac{1}{4}]$ by the 
second agent.
The second agent then marks and is
assigned the slice $[\frac{1}{4},\frac{1}{2}]$.
The third agent then marks and is
assigned the slice $[\frac{1}{2},\frac{3}{4}]$,
leaving the fourth agent with the
slice $[\frac{3}{4},1]$. 
The procedure is not proportional as the
fourth agent only receives $\frac{1}{6}$ of
the total value, 
not (weakly) envy free as the
first agent envies the fourth agent, 
and not equitable as agents receive
cake of different value.
The procedure is not (weakly) Pareto optimal
as allocating the first agent
with $[\frac{3}{4},1]$, the second 
with $[\frac{1}{2},\frac{3}{4}]$,
the third with $[0,\frac{1}{4}]$,
and the fourth with $[\frac{1}{4},\frac{1}{2}]$
gives all agents greater 
value. 

The procedure is not truthful
as the second agent can 
get a larger and more valuable 
slice by misrepresenting their preferences
and marking the cake into
the slices $[\frac{1}{4},\frac{5}{8}]$,
$[\frac{5}{8},\frac{3}{4}]$,
and $[\frac{3}{4},1]$. 
In this situation, the third agent allocates
the second agent with the slice
$[\frac{1}{4},\frac{5}{8}]$ which is of
greater value to the second agent.

Finally, to show that the procedure is not order
monotonic consider three agents
and a cake in which
the first agent places equal value
on each of $[0,\frac{1}{3}]$,
$[\frac{1}{3},\frac{2}{3}]$
and $[\frac{2}{3},1]$, 
the second 
places no value on $[0,\frac{1}{3}]$,
half the total value on $[\frac{1}{3},\frac{2}{3}]$,
and one quarter on each of $[\frac{2}{3},\frac{5}{6}]$,
and $[\frac{5}{6},1]$, 
and the third 
places a value of 
one sixth the total value on $[0,\frac{1}{6}]$,
no value on $[\frac{1}{6},\frac{1}{3}]$
and $[\frac{1}{3},\frac{2}{3}]$, 
and half the remaining value on $[\frac{2}{3},\frac{5}{6}]$
and $[\frac{5}{6},1]$. 
The first agent marks and is 
allocated the slice $[0,\frac{1}{3}]$.
The second agent marks
and is allocated the slice $[\frac{1}{3},\frac{2}{3}]$,
leaving the third agent with the slice
$[\frac{2}{3},1]$. On the other hand, 
suppose the third agent arrives ahead
of the second agent. In this 
case, the third agent marks 
the cake into two slice, $[\frac{1}{3},\frac{5}{6}]$
and $[\frac{5}{6},1]$. The second agent
allocates the third agent the slice $[\frac{5}{6},1]$.
Hence, the value of the interval allocated to the third agent
halves when they go
second in the arrival order. 
Hence the procedure is not order
monotonic. 
\myqed

\section{Related work}

There is an extensive literature on fair
division and cake cutting procedures. 
See, for instance, 
\cite{cakecut}. There has, however,
been considerably less work on
fair division problems similar 
to those considered here. 
Thomson considers a generalization
where the number of agents may increase
\cite{wmor83}.  He explores
whether it is possible to have
a procedure in which agents'
allocations are monotonic (i.e.
their values do not increase as
the number of agents increase)
combined with other common properties like
weak Pareto optimality. 
Cloutier {\it et al.} consider
a different generalization of 
the cake cutting problem in which
the number of agents is fixed
but there are multiple cakes
\cite{cmsc2010}. This models
situations where, for example, agents
wish to choose shifts across
multiple days. This problem
cannot be reduced to multiple single
cake cutting problems if the agents'
valuations across cakes are linked
(e.g. you prefer the same shift each day). 
%
A number of authors have studied
distributed mechanisms for fair
division (see, for example,
\cite{cem09}).
In such mechanisms, agents typically agree locally on deals to exchange
goods. 
The usual goal is to identify conditions under which
the system converges to
a fair or envy free allocation.

\section{Conclusions}

We have proposed an online form of the 
cake cutting problem. 
This permits us to explore
the concept of fair division when
agents arrive and depart during the
process of dividing a resource. 
It can be used to model situations,
such as on the internet, when we need
to divide resources asynchronously. 
There are many possible future directions for this
work. One extension
would be to undesirable goods (like chores)
where we want as little of them as possible. 
It would also be interesting to
consider the variation of the
problem where agents
have partial information about the
valuation functions of the
other agents. For voting and other forms
of preference aggregation, there has been
considerable interest of late in reasoning about
preferences that are incomplete or partially
known \cite{prvwijcai2007,waaai2007,prvwkr08}. 
With cake cutting, agents can act
more strategically when they have such partial
knowledge.

\section*{Acknowledgments}

Toby Walsh 
is supported by the Australian Department 
of Broadband, Communications and the Digital Economy,
the ARC, and the Asian Office of Aerospace 
Research and Development (AOARD-104123).

%

\myOmit{

}

\end{document}